\documentclass[10pt,twocolumn,letterpaper]{article}

\usepackage{iccv}
\usepackage{times}
\usepackage{epsfig}
\usepackage{graphicx}
\usepackage{amsmath}
\usepackage{amssymb}
\usepackage{wrapfig}
\usepackage{subcaption}
\usepackage{authblk}
\newcommand{\nothing}[1]{}

\usepackage[pagebackref=true,breaklinks=true,letterpaper=true,colorlinks,bookmarks=false]{hyperref}

\iccvfinalcopy 

\def\iccvPaperID{3593} 

\ificcvfinal\fi
\begin{document}

\title{Learning Perspective Undistortion of Portraits}
\author[1,*]{Yajie Zhao}
\author[1,2,*]{Zeng Huang}
\author[1,2]{Tianye Li}
\author[1]{Weikai Chen}
\author[1,2]{Chloe LeGendre}
\author[1]{Xinglei Ren}
\author[1]{Jun Xing}
\author[1]{Ari Shapiro}
\author[1,2,3]{Hao Li}
\affil[1]{USC Institute for Creative Technologies}
\affil[2]{University of Southern California}
\affil[3]{Pinscreen}

\twocolumn[{%
\renewcommand\twocolumn[1][]{#1}%
\maketitle
\vspace{-0.3in}
\begin{center}
    \centering
   \vspace{-14pt}
  \includegraphics[width=\linewidth]{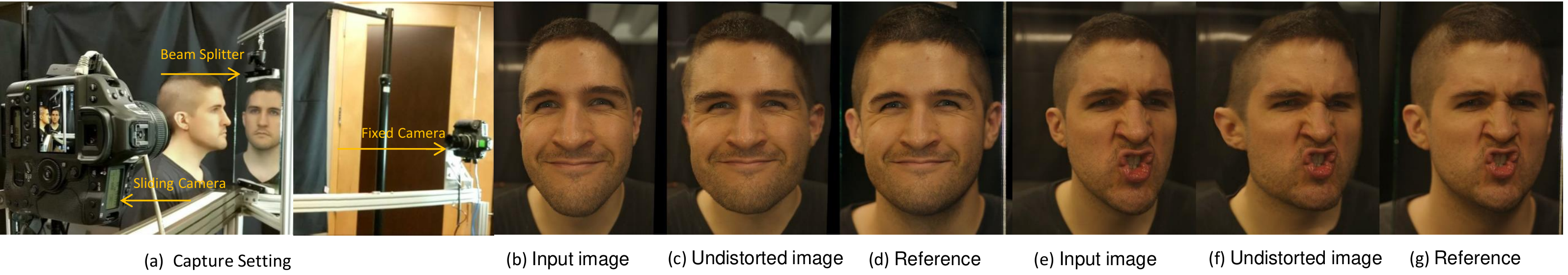}\\
   \captionof{figure}{\small{We propose a learning-based method to remove perspective distortion from portraits. For a subject with two different facial expressions, we show input photos (b) (e), our undistortion results (c) (f), and reference images (d) (g) captured simultaneously using a beam splitter rig (a). Our approach handles even extreme perspective distortions.}}
    \label{fig:teaser}
\vspace{0.05in}
\end{center}
}]

\begin{abstract}
Near-range portrait photographs often contain perspective distortion artifacts that bias human perception and challenge both facial recognition and reconstruction techniques.
We present the first deep learning based approach to remove such artifacts from unconstrained portraits. In contrast to the previous state-of-the-art approach \cite{fried2016perspective}, our method handles even portraits with extreme perspective distortion, as we avoid the inaccurate and error-prone step of first fitting a 3D face model. Instead, we predict a distortion correction flow map that encodes a per-pixel displacement that removes distortion artifacts when applied to the input image. Our method also automatically infers missing facial features, i.e. occluded ears caused by strong perspective distortion, with coherent details.
We demonstrate that our approach significantly outperforms the previous state-of-the-art \cite{fried2016perspective} both qualitatively and quantitatively, particularly for portraits with extreme perspective distortion or facial expressions. We further show that our technique benefits a number of fundamental tasks, significantly improving the accuracy of both face recognition and 3D reconstruction and enables a novel camera calibration technique from a single portrait. Moreover,
we also build the first perspective portrait database
with a large diversity in identities, expression and poses.
\end{abstract}

\vspace{-10pt}
\section{Introduction}
\nothing{
\begin{itemize}
    \item The near-range selfie shooting tends to distort facial features, e.g. enlarged nose, puzzling people in social media communication.
    \item The facial distortion leads to inaccurate 3D face reconstruction.
    \item Face recognition can be impaired by perspective transformation. \cite{liu2003face,liu2006face}
\end{itemize}
}

Perspective distortion artifacts are often observed in portrait photographs, in part due to the popularity of the ``selfie" image captured at a near-range distance. The inset images, where a person is photographed from distances of 160$cm$ and 25$cm$, demonstrate these artifacts. \begin{wrapfigure}{l}{0.22\textwidth}
	\vspace*{-11mm}
	\hspace*{-2.2mm}
	\begin{center}
		\includegraphics[width=0.25\textwidth]{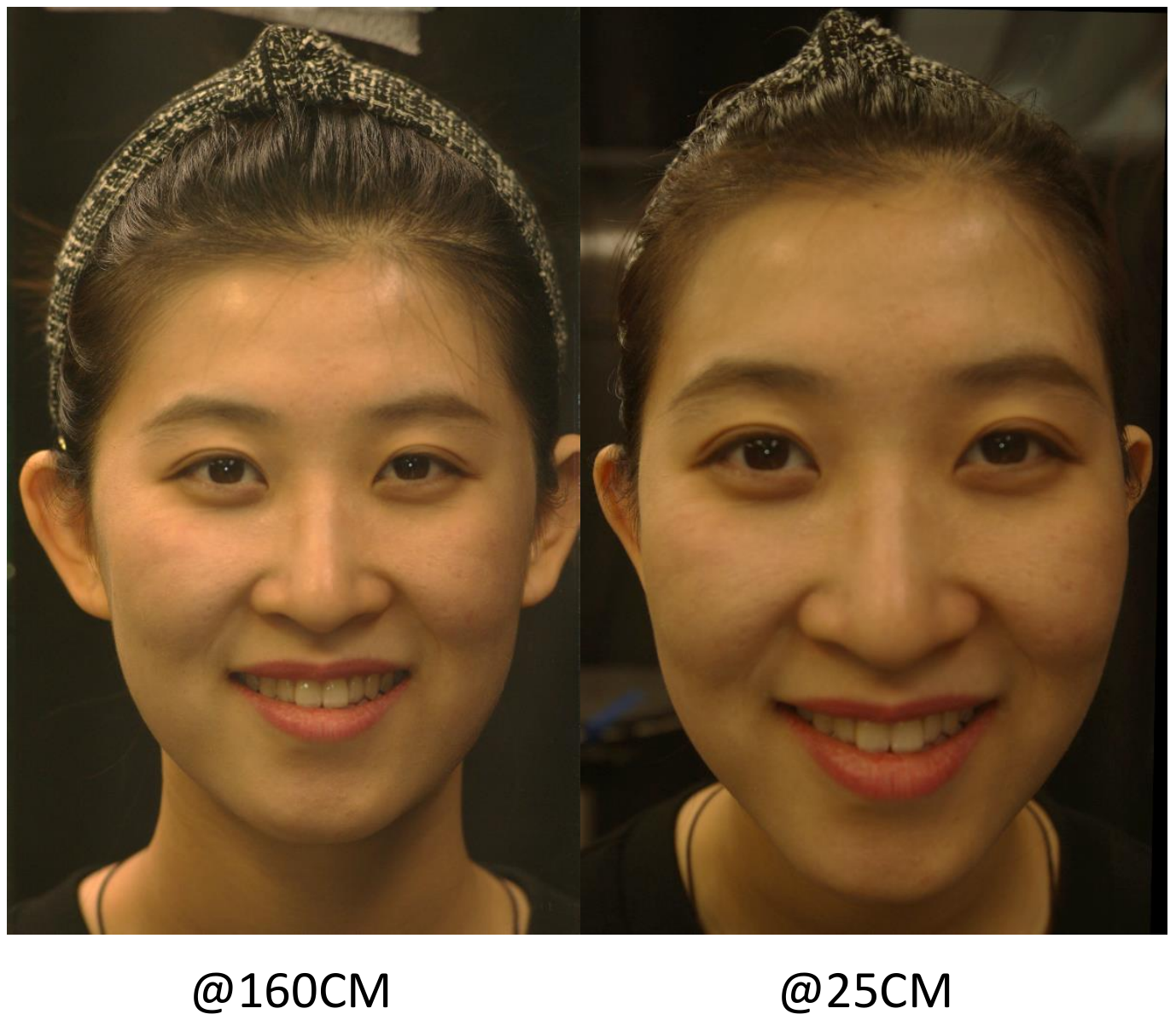}
	\end{center}
	\label{fig:twoviews}
	\vspace*{-6.8mm}
\end{wrapfigure} When the object-to-camera distance is comparable to the size of a human head, as in the 25$cm$ distance example, there is a large proportional difference between the camera-to-nose distance and camera-to-ear distance. This difference creates a face with unusual proportions, with the nose and eyes appearing larger and the ears vanishing all together \cite{ward2018nasal}.

Perspective distortion in portraits not only influences the way humans perceive one another \cite{bryan2012perspective}, but also greatly impairs a number of computer vision-based tasks, such as face verification and landmark detection. Prior research \cite{liu2003face,liu2006face} has demonstrated that face recognition is strongly compromised by the perspective distortion of facial features. Additionally, 3D face reconstruction from such portraits is highly inaccurate, as geometry fitting starts from biased facial landmarks and distorted textures.  

Correcting perspective distortion in portrait photography is a largely unstudied topic. Recently, Fried et al.~\cite{fried2016perspective} investigated a related problem, aiming to manipulate the relative pose and distance between a camera and a subject in a given portrait. Towards this goal, they fit a full perspective camera and parametric 3D head model to the input image and performed 2D warping according to the desired change in 3D. However, the technique relied on a potentially inaccurate 3D reconstruction from facial landmarks biased by perspective distortion. Furthermore, if the 3D face model fitting step failed, as it could for extreme perspective distortion or dramatic facial expressions, so would their broader pose manipulation method. In contrast, our approach does not rely on model fitting from distorted inputs and thus can handle even these challenging inputs. Our GAN-based synthesis approach also enables high-fidelity inference of any occluded features, not considered by Fried et al. ~\cite{fried2016perspective}.

In our approach, we propose a cascaded network that maps a near-range portrait with perspective distortion to its distortion-free counterpart at a canonical distance of $1.6m$ (although any distance between $1.4m \sim 2m$ could be used as the target distance for good portraits). Our cascaded network includes a {distortion correction flow} network and a {completion} network. Our {distortion correction flow} method encodes a per-pixel displacement, maintaining the original image's resolution and its high frequency details in the output. However, as near-range portraits often suffer from significant perspective occlusions, flowing individual pixels often does not yield a complete final image. Thus, the {completion} network inpaints any missing features. A final texture blending step combines the face from the completion network and the warped output from the distortion correction network. As the possible range of per-pixel flow values vary by camera distance, we first train a {camera distance prediction} network, and feed this prediction along with the input portrait to the distortion correction network.

\nothing{We propose to learn a high-dimensional embedding that maps a portrait with perspective distortion to a distortion-free counterpart in the same semantic. In order to undistort the unconstrained portraits in arbitrary resolution, we define a \textit{distortion correction flow map} that encodes a per-pixel displacement with the input, such that when the flow map is applied to the input portrait, its perspective distortion artifacts are removed with the output in the original resolution. While our perspective distortion correction flow mapping can faithfully remove many artifacts, near-portrait photographs often suffer from significant perspective-related occlusions, e.g. missing ears or failing to capture the outer rim of the face. Thus, a simple mapping from the distorted input to the undistorted output results in holes due to such occlusions. We therefore propose an cascade image completion network that further inpaints occluded facial content with coherent details. A blending step is further applied to combine complete face from inpainting network and output warped from flow. }

Training our proposed networks requires a large corpus of paired portraits with and without perspective distortion. However, to the best of our knowledge, no previously existing dataset is suited for this task. As such, we construct the first portrait dataset rendered from 3D head models with large variations in camera distance, head pose, subject identity, and illumination. To visually and numerically evaluate the effectiveness of our approach on real portrait photographs, we also design a beam-splitter photography system (see Teaser) to capture portrait pairs of real subjects simultaneously on the same optical axis, eliminating differences in poses, expressions and lighting conditions.

Experimental results demonstrate that our approach removes a wide range of perspective distortion artifacts (e.g., increased nose size,  squeezed face, {\it etc}), and even restores missing facial features like ears or the rim of the face. We show that our approach significantly outperforms Fried et al.~\cite{fried2016perspective} both qualitatively and quantitatively for a synthetic dataset, constrained portraits, and unconstrained portraits from the Internet. We also show that our proposed face undistortion technique, when applied as a pre-processing step, improves a wide range of fundamental tasks in computer vision and computer graphics, including face recognition/verification, landmark detection on near-range portraits (such as head mounted cameras in visual effects), and 3D face model reconstruction, which can help 3D avatar creation and the generation of 3D photos (Section~\ref{sec:application}). Additionally, our novel \textit{camera distance prediction} provides accurate camera calibration from a single portrait. 

Our main contributions can be summarized as follows:
\begin{itemize}
    \item The first deep learning based method to automatically remove perspective distortion from an unconstrained near-range portrait, benefiting a wide range of fundamental tasks in computer vision and graphics.
    \item A novel and accurate camera calibration approach that only requires a single near-range portrait as input.
    \item A new perspective portrait database for face undistortion with a wide range of subject identities, head poses, camera distances, and lighting conditions. 
\end{itemize}

\section{Related Work}
\label{sec:related_work}

\paragraph{Face Modeling.} We refer the reader to \cite{parke2008computer} for a comprehensive overview and introduction to the modeling of digital faces. With advances in 3D scanning and sensing technologies, sophisticated laboratory capture systems \cite{beeler2010high, beeler2011high, bradley2010high, debevec2000acquiring, ghosh2011multiview, li2009robust, ma2007rapid, weise2009face} have been developed for high-quality face reconstruction. However, 3D face geometry reconstruction from a single unconstrained image remains challenging. The seminal work of Blanz and Vetter~\cite{blanz1999morphable} proposed a PCA-based morphable model, which laid the foundation for modern image-based 3D face modeling and inspired numerous extensions including face modeling from internet pictures~\cite{kemelmacher2013internet}, multi-view stereo \cite{amberg2007reconstructing}, and reconstruction based on shading cues \cite{kemelmacher20113d}. To better capture a variety of identities and facial expressions, the multi-linear face models \cite{vlasic2005face} and the FACS-based blendshapes \cite{cao2014facewarehouse} were later proposed. When reconstructing a 3D face from images, sparse 2D facial landmarks \cite{cootes2001active, cristinacce2008automatic, saragih2011deformable, xiong2013supervised} are widely used for a robust initialization. Shape regressions have been exploited in the state-of-the-art landmark detection approaches \cite{cao2014face, kazemi2014one, ren2014face} to achieve impressive accuracy.

Due to the low dimensionality and effectiveness of morphable models in representing facial geometry, there have been significant recent advances in single-view face reconstruction \cite{thies2016face2face, richardson2017learning, kim2017inversefacenet, tewari2017mofa, hu2017avatar}. However, for near-range portrait photos, the perspective distortion of facial features may lead to erroneous reconstructions even when using the state-of-the-art techniques. Therefore, portrait perspective undistortion must be considered as a part of a pipeline for accurately modeling facial geometry.

\paragraph{Face Normalization.}
Unconstrained photographs often include occlusions, non-frontal views, perspective distortion, and even extreme poses, which introduce a myriad of challenges for face recognition and reconstruction. However, many prior works \cite{hassner2015effective,yin2017towards,sagonas2015robust,huang2017beyond} only focused on normalizing head pose. Hassner et al.~\cite{hassner2015effective} ``frontalized" a face from an input image by estimating the intrinsic camera matrix given a fixed 3D template model. Cole et al.~\cite{cole2017synthesizing} introduced a neural network that mapped an unconstrained facial image to a front-facing image with a neutral expression. Huang et al.~\cite{huang2017beyond} used a generative model to synthesize an identity-preserving frontal view from a profile. Bas et al.~\cite{bas2017does} proposed an approach for fitting a 3D morphable model to 2D landmarks or contours under either orthographic or perspective projection. 

Psychological research suggests a direct connection between camera distance and human portrait perception. Bryan et al.~\cite{bryan2012perspective} showed that there is an ``optimal distance" at which portraits should be taken. Cooper et al.~\cite{cooper2012perceptual} showed that the 50mm lens is most suitable for photographing an undistorted facial image. Valente et al.~\cite{valente2015perspective} proposed to model perspective distortion as a one-parameter family of warping functions with known focal length. Most related to our work, Fried et al.~\cite{fried2016perspective} investigated the problem of editing the facial appearance by manipulating the distance between a virtual camera and a reconstructed head model. Though this technique corrected some mild perspective distortion, it was not designed to handle extreme distortions, as it relied on a 3D face fitting step. In contrast, our technique requires no shape prior and therefore can generate undistorted facial images even from highly distorted inputs.

\paragraph{Image-based Camera Calibration.} 
Camera calibration is an essential prerequisite for extracting precise and reliable 3D metric information from images. We refer the reader to \cite{remondino2006digital, salvi2002comparative, medioni2004emerging} for a survey of such techniques. The state-of-the-art calibration methods mainly require a physical target such as checkerboard pattern \cite{tsai1987versatile,zhang2000flexible} or circular control points \cite{chen2004camera, colombo2006camera, heikkila2000geometric,jiang2005detection, datta2009accurate}, used for locating point correspondences. Flores et al.~\cite{flores2013camera} proposed the first method to infer camera-to-subject distance from a single image with a calibrated camera. Burgos-Artizzu et al.~\cite{burgos2014distance} built the Caltech Multi-Distance Portraits Dataset (CMDP) of portraits of a variety of subjects captured from seven distances. Many recent works directly estimate camera parameters using deep neural networks. PoseNet~\cite{kendall2015posenet} proposed an end-to-end solution for 6-DOF camera pose localization. Others ~\cite{workman2015deepfocal, workman2016hlw,hold2018perceptual} proposed to extract camera parameters using vanishing points from a single scene photos with horizontal lines. To the best of our knowledge, our method is the first to estimate camera parameters from a single portrait.




\section{Overview}
\label{sec:pipeline}
\begin{figure*}[htb]
\begin{center} 
   \includegraphics[width=0.9\linewidth]{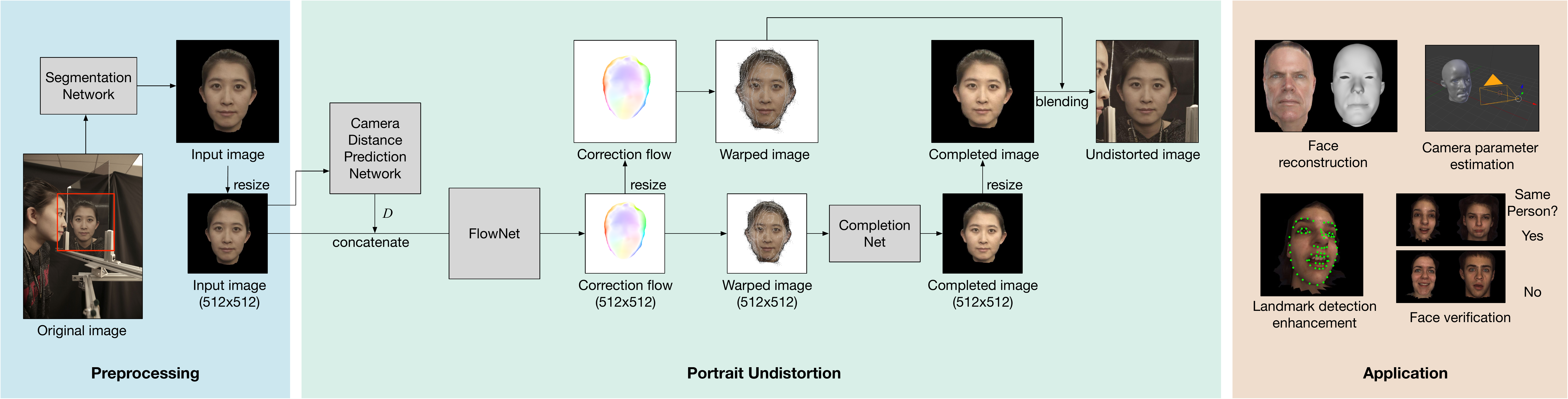}
   \end{center}
  \caption{The pipeline workflow and applications of our approach. The input portrait is first segmented and scaled in the preprocessing stage and then fed to a network consisting of three cascaded components. The \textit{FlowNet} rectifies the distorted artifacts in the visible regions of input by predicting a distortion correction flow map. The \textit{CompletionNet} inpaints the missing facial features due to the strong perspective distortions and obtains the completed image. The outcomes of two networks are then scaled back to the original resolution and blended with high-fidelity mean texture to restore fine details.}

  \label{fig:pipeline}
\end{figure*}
\begin{figure}[htb]
  \centering 
   \includegraphics[width=1.0\linewidth]{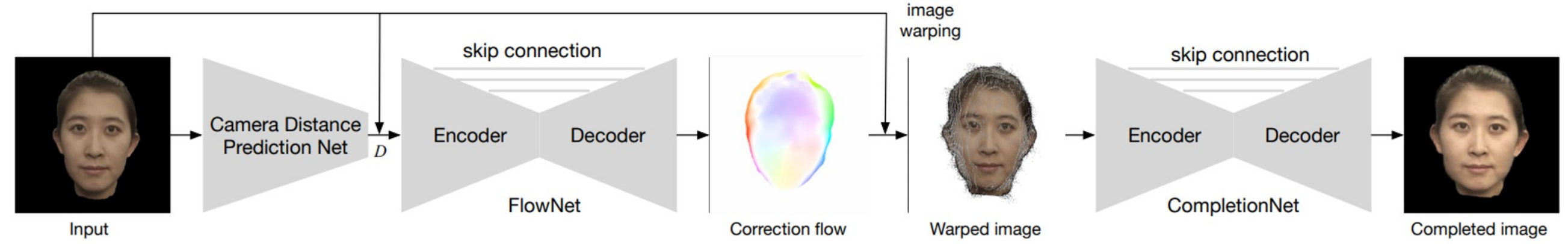}
  \caption{Cascade Network Structure.}
  {
}
  \label{fig:network}
\end{figure}

The overview of our system is shown in Fig.~\ref{fig:pipeline}. We pre-process the input portraits with background segmentation, scaling, and spatial alignment (see appendix), and then feed them to a \textit{camera distance prediction} network to estimates camera-to-subject distance. The estimated distance and the portrait are fed into our cascaded network including \textit{FlowNet}, which predicts a distortion correction flow map, and \textit{CompletionNet}, which inpaints any missing facial features caused by perspective distortion. Perspective undistortion is not a typical image-to-image translation problem, because the input and output pixels are not spatially corresponded. Thus, we factor this challenging problem into two sub tasks: first finding a per-pixel undistortion flow map, and then image completion via inpainting. In particular, the vectorized flow representation undistorts an input image at its original resolution, preserving its high frequency details, which would be challenging if using only generative image synthesis techniques. In our cascaded architecture (Fig.~\ref{fig:network}), \textit{CompletionNet} is fed the warping result of \textit{FlowNet}. We provide details of \textit{FlowNet} and \textit{CompletionNet} in Sec.~\ref{sec:faceUndistort}. Finally, we combined the results of the two cascaded networks using the Laplacian blending~\cite{adelson1984pyramid} to synthesize high-resolution details while maintaining plausible blending with existing skin texture.

\section{Portrait Undistortion}
\label{sec:faceUndistort}

\begin{figure}[t]
  \centering
\includegraphics[width=0.85\linewidth]{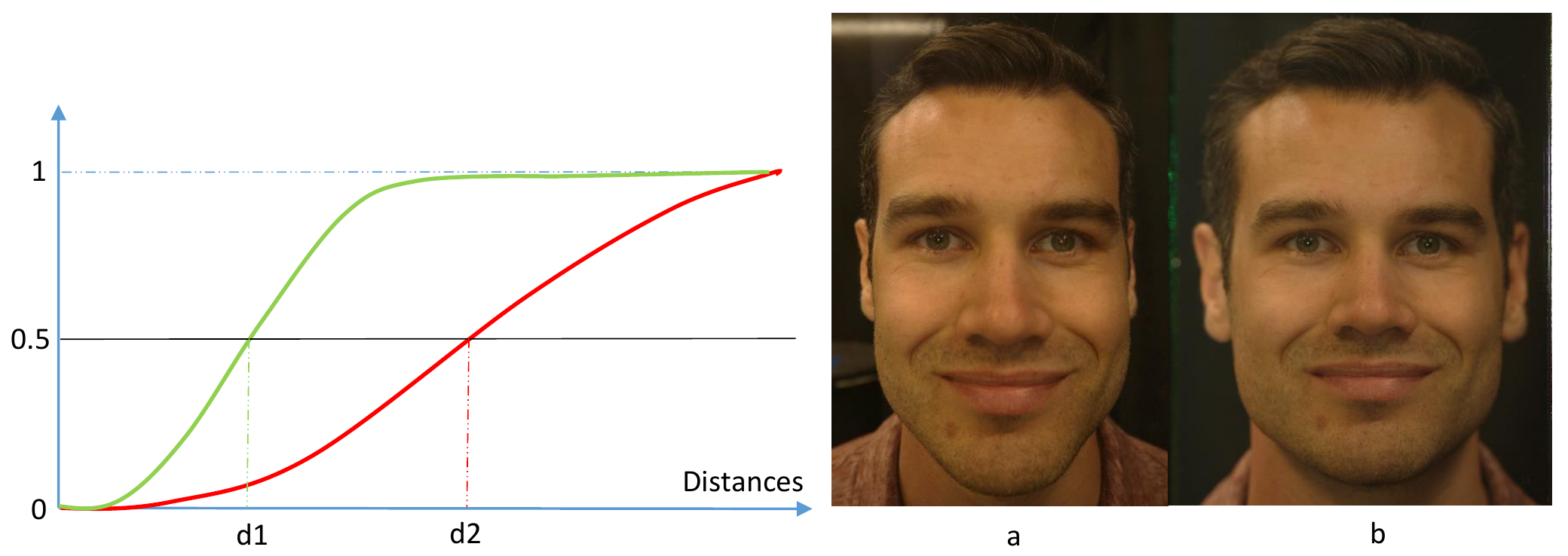}
  \caption{Illustration of Camera Distance Prediction Classifier. \textbf{\textit{Green Curve}} and \textbf{\textit{Red Curve}} are the response curves of input \textbf{\textit{a}} and \textbf{\textit{b}}; \textbf{\textit{d1}} and \textbf{\textit{d2}} are the predicted distances of input \textbf{\textit{a}} and input \textbf{\textit{b}}.} 
  \label{fig:distanceClassifier}
\end{figure}

\subsection{Camera Distance Prediction Network} 
\label{sec:camDistance}
Rather than regress directly from the input image to the camera distance $D$, which is known to be challenging to train, we use a distance classifier. We check if the camera-to-subject distance of the input is larger than an query distance $d$ $\epsilon$ (17.4cm, 130cm)\footnote{17.4cm and 130cm camera-to-subject distances correspond to 14mm and 105mm in 35mm equivalent focal length}. Our strategy learns a continuous mapping from input images to the target distances. Given any query distance $D$ and input image pair $(input, D)$, the output is a floating point number in the range of $0 \sim 1$, which indicates the probability that the distance of the input image is greater than the query distance $D$. As shown in Fig.~\ref{fig:distanceClassifier}, the vertical axis indicates the output of our distance prediction network while the horizontal axis is the query distance. To predict the distance, we locate the query distance with a network output of 0.5. With our network denoted as $\mathcal{\phi}$, our network holds the transitivity property that if $d1 > d2$, $\mathcal {\phi}(input, d1) > \mathcal{\phi}(input, d2)$.

To train the camera distance network, we append the value of $\log_2{d}$ as an additional channel of the input image and extract features from the input images using the VGG-11 network~\cite{simonyan2014very}, followed by a classic classifier consisting of fully connected layers. As training data, for each of the training image with ground truth distance $d$, we sample a set of $\log_2{d}$ using normal distribution $\log_2{d} \sim \mathcal{N}(\log_2{D} ,0.5^2)$.

\subsection{FlowNet}
\label{sec:flownet}

The FlowNet operates on the normalized input image ($512 \times 512$) $\mathcal{A}$ and estimates a correction forward flow $\mathcal{F}$ that rectifies the facial distortions.
However, due to the immense range of possible perspective distortions, the correction displacement for portraits taken at different distances will exhibit different distributions.
Directly predicting such high-dimensional per-pixel displacement is highly under-constrained and often leads to inferior results (Figure~\ref{fig:ablationWithLabel}). 
To ensure more efficient learning, we propose to attach the estimated distance to the input of FlowNet in the similar way as in Section~\ref{sec:camDistance}. Instead of directly attaching the predicted number, we propose to classify these distances into eight intervals\footnote{The eight distance intervals are $[23, 26)$, $[26, 30)$, $[30, 35)$, $[35, 43)$, $[43, 62)$, $[62, 105)$, $[105, 168)$ and $[168, +\infty )$, which are measured in centimeters.} and use the class label as the input to \textit{FlowNet}. The use of label will decrease the risk of accumulation error from camera distance prediction network, because the accuracy of predicting a label is higher than floating number. 

\textit{FlowNet} takes $\mathcal{A}$ and distance label $\mathcal{L}$ as input, and it will predict a forward flow map $\mathcal{F_{AB}}$, which can be used to obtain undistorted output $\mathcal{B}$ when applied to $\mathcal{A}$. For each pixel $(x, y)$ of $\mathcal{A}$, $\mathcal{F_{AB}}$ encodes the translation vector $(\Delta_x, \Delta_y)$. Denote the correspondence of $(x, y)$ on $\mathcal{B}$ as $(x', y')$, then $ (x', y')  = (x, y)  + (\Delta_x, \Delta_y) $. In \textit{FlowNet}, we denote generator and discriminator as $G$ and $D$ separately. Then L1 loss of flow is as below:

\begin{equation}
\mathcal{L}_{G} = \| \boldsymbol{y} - \mathcal{F_{AB}} \|_1
\end{equation}
In which $\boldsymbol{y}$ is the ground truth flow. For the discriminator loss, as forward flow $\mathcal{F_{AB}}$ is per-pixel correspondence to $\mathcal{A}$ but not $\mathcal{B}$, thus $\mathcal{B}$ will have holes, seams and discontinuities which is hard to used in discriminator. To make the problem more tractable, instead of applying discriminator on $\mathcal{B}$, we use the $\mathcal{F_{AB}}$ to map $\mathcal{B}$ to $\mathcal{A'}$ and use $\mathcal{A}$ and $\mathcal{A'}$ as pairs for discriminator on the condition of $\mathcal{L}$.

\begin{equation}
\resizebox{0.7\hsize}{!}{$
\begin{split}
\mathcal{L}_{\mathrm{D}} & = \min _{G} \max _{D}\
\mathbb{E}_{\boldsymbol{x}
	\sim
	p_\mathrm{data}(\boldsymbol{x})} \big[ \mathrm{log}\
D(\boldsymbol{\mathcal{A}}, \boldsymbol{\mathcal{L}}) \big] + \\ &\mathbb{E}_{\boldsymbol{z} \sim
	p_{\boldsymbol{z}}(\boldsymbol{z})}\big[ \mathrm{log}\ (1 -
D(\mathcal{A'}, \boldsymbol{\mathcal{L}}))\big].
\end{split}
$}
\end{equation}

where $p_\mathrm{data}(\boldsymbol{x})$ and $p_{\boldsymbol{z}}(\boldsymbol{z})$ represent the distributions of real data $\boldsymbol{x}$(input image  $\mathcal{A}$ domain) and noise variables $\boldsymbol{z}$ in the domain of $\mathcal{A}$ respectively. The discriminator will penalize the joint configuration in the space of $\mathcal{A}$, which leads to shaper results.

\subsection{CompletionNet}
\label{sec:completionNet}

The distortion-free result will then be fed into the CompletionNet, which focuses on inpainting missing features and filling the holes. 
Note that as trained on a vast number of paired examples with large variations of camera distance, CompletionNet has learned an adaptive mapping regarding to varying distortion magnitude inputs. 
\nothing{In particular, an input image with complete facial features will remain largely unchanged after passed through the network while the ones that still lack facial clues will be automatically completed.}


\subsection{Network Architecture}
\label{sec:architecture}

\nothing{Both the CompletionNet and the sub-networks in FlowNet are implemented using identical architectures.
In particular, we employ a U-net structure with skip connections similar to \cite{isola2017image}.
Such architecture is well-suited to our task, as the skip connections between layers of the encoder and decoder modules allow for easy preservation of the overall structure of the input in the output image while avoiding the artifacts and limited resolutions found in more typical encoder-decoder networks.
This enables the network to leverage more of its overall capacity to learn appropriate transformation from the provided input to the desired output.
As we perform inference directly in the image space, 
In addition, the face completion task is likely to introduce significant topology changes when the missing features are recovered.
We found that trivially applying the U-net structure fails to generate reasonable results for our purpose.
To make the problem more tractable, we introduce several modifications to improve the resulting image quality and stabilize the training process.
The details are stated below.}

We employ a U-net structure with skip connections similar to \cite{isola2017image} to both \textit{FlowNet} and \textit{CompletionNet}. There is not direct correspondence between each pixel in the input and those in the output. In the \textit{FlowNet}, the L1 and GAN discriminator loss are only computed within the segmentation mask, leading the network to focus on correcting details that only will be used in the final output. In the \textit{CompletionNet}, as the output tends to cover more pixels than the input, during training, we compute a new mask that denotes the novel region compared to input and assign higher L1 loss weight to this region.
In implementation, we set the weight ratio of losses computed inside and outside the mask as $5:1$ for the \textit{CompletionNet}.
We found that this modifications leads to better inference performance while producing results with more details.

\paragraph{Implementation Details.}
We train \textit{FlowNet} and \textit{CompletionNet} separately using the Adam optimizer \cite{kingma2014adam} with learning rate 0.0002.
All training was performed on an NVIDIA GTX 1080 Ti graphics card.
For both networks, we set the weights of L1 and GAN loss as 10.0 and 1.0 respectively. 
As shown in Fig.~\ref{fig:network}, the generator in both networks uses the mirrored structure, in which both encoder and decoder use 8-layer convolutional network. 
ReLU activation and batch normalization are used in all layers except the input and output layers.
The discriminator consists of four convolutional layers and one fully connected layer.



\section{Data Preparation}
\label{sec:data}
\nothing{A neural network based approach would na\"ively require pairs of portraits with and without the perspective distortion, ideally including diverse subjects, expressions, illuminations and head poses from a variety of distances. Although the Caltech Multi-Distance Portrait Dataset (CMDP) \cite{burgos2014distance} provides datasets of 53 subjects captured at 7 distances, as these images were not captured simultaneously, both camera pose and facial expressions change slightly across distances for each subject, which cannot define as a pair with only perspective changing. We therefore generate a novel training dataset where we can control the subject-to-camera distance, head pose, and illumination and ensure that the image differences are only caused \textit{only} by perspective distortion.
}

\paragraph{Training Data Acquisition and Rendering.}As there is no database of paired portraits with \textit{only} perspective changing. We therefore generate a novel training dataset where we can control the subject-to-camera distance, head pose, illumination and ensure that the image differences are only caused \textit{only} by perspective distortion. Our synthetic training corpus is rendered from 3D head models acquired by two scanning system. The first is a light-stage\cite{ghosh2011multiview, ma2007rapid} scanning system which produces pore-level 3D head models for photo-real rendering. Limited by the post-processing time, cost and number of individuals that can be rapidly captured, we also employ a second capture system engineered for rapid throughput. In total, we captured 15 subjects with well defined expressions in the high-fidelity system, generating 307 individual meshes, and 200 additional subjects in the rapid-throughput system.
\nothing{Since we have the 3D ground-truth face geometry and the corresponding reflectance maps, during rendering, it is easy to trace the 2D projected pixels of the same surface point and thereby compute the flow maps for network training and evaluation. } 

We rendered synthetic portraits using a variety of camera distances, head poses, and incident illumination conditions. We sample distances distributed from 23$cm$ to 160$cm$ with corresponding 35$mm$ equivalent focal length from 18$mm$ to 128$mm$ to ensure the same framing. We also randomly sample candidate head poses in the range of -45$^\circ$ to +45$^\circ$ in pitch, yaw, and roll. We used 107 different image-based environment for global illumination combined with point lights. With the random combination of camera distances, head poses, and illuminations, we totally generate 35,654 pairs of distroted/ undistored portraits along with forward flow. \nothing{we generate 18,164 and 17,490 pairs of distroted/ undistored portraits using two scanning system separately. We render all the portraits using a OPENGL skin shader under perspective projection model with SSS (?) tecnquies in real-time. We also generate the forward flow by compute the pixel displacement between distorted view and undistorted view. }

17,000 additional portrait pairs warped from U-4DFE dataset \cite{yin2008high}(38 females and 25 males subjects for training while 20 females and 17 males for testing) are used to expand diversity of identities.

\begin{figure}
\begin{center}
   \includegraphics[width=1\linewidth]{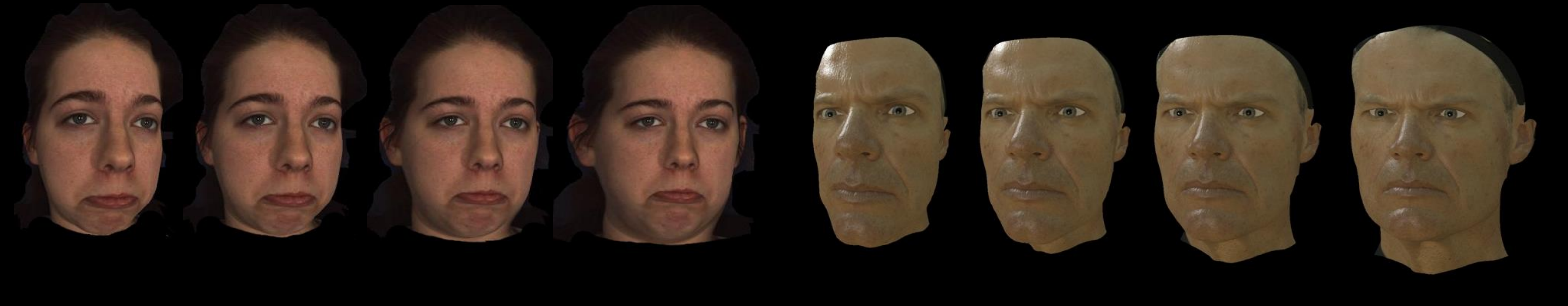}
\end{center}
   \caption{Training dataset. \textit{\textbf{The left side triplet}} are the synthetic images generated from BU-4DFE dataset\cite{yin2008high}. \textit{\textbf{From the left to the right}}, the camera-to-subject distances are: \textit{24cm, 30cm, 52cm, 160cm}. \textit{\textbf{The right side triplet}} are the synthetic images rendered from high-resolution 3D model. \textit{\textbf{From left to the right}}, the camera-to-subject distances are: \textit{22cm, 30cm, 53cm, 160cm}. }
\label{fig:dataset}
\end{figure}

\nothing{
and real photos,
To supplement this data with additional identities and real photos, we also generate portraits on the 3D facials scans of the BU-4DFE dataset \cite{yin2008high} \nothing{after randomly sampling candidate head poses in the range of -10$^\circ$ to +10$^\circ$ in pitch and yaw. } Out of 58 female subjects and 43 male subjects total, we trained using 38 female and 25 male subjects, keeping the rest set aside as test data, for a total of 17,000 additional training pairs. Fig.~\ref{fig:dataset} show the examples of our training data.}

\paragraph{Test Data Acquisition.} To demonstrate that our system scales well to real-world portraits, we also devised a two-camera beam splitter capture system that would enable simultaneous photography of a subject at two different distances. As shown in Teaser left, we setup a beam-splitter (50\% reflection, 50\% transmission) at 45$^\circ$ along a metal rail, such that a first DSLR camera was set at the canonical fixed distance of 1.6$m$ from the subject, with a 180$mm$ prime lens to image the subject directly, while a second DSLR camera was set at a variable distance of 23$cm$ - 1.37$m$, to image the subject's reflection off the beam-splitter with a 28$mm$ zoom lens.
With carefully geometry and color calibration, the two hardware-synced cameras were able to capture nearly ground truth portraits pairs of real subjects both with and without perspective distortions. (\textit{More details can be found in the appendix}).

\nothing{Before feeding an input photo to our framework, we first detect its face bounding box using the method of \cite{viola2004robust} and then segment out the head region with a segmentation network in\cite{long2015fully}. The network is trained with modified portrait data from \cite{shen2016automatic}. The input is further scaled and aligned with the training data format.}

\nothing{In order to handle images with arbitrary resolution, we pre-process the segmented images to a uniform size of $512 \times 512$. The input image is first scaled so that its detected inter-pupillary distance matches a target length, computed by averaging that of ground truth images. We then crop the image to $512 \times 512$, maintaining the right eye inner alignment. Images too small to be cropped are padded with black.}

\section{Results and Experiments}
\label{sec:result}

\subsection{Evaluations}
\label{sec:evaluation}

\begin{figure}
\begin{center}
   \includegraphics[width=0.9\linewidth]{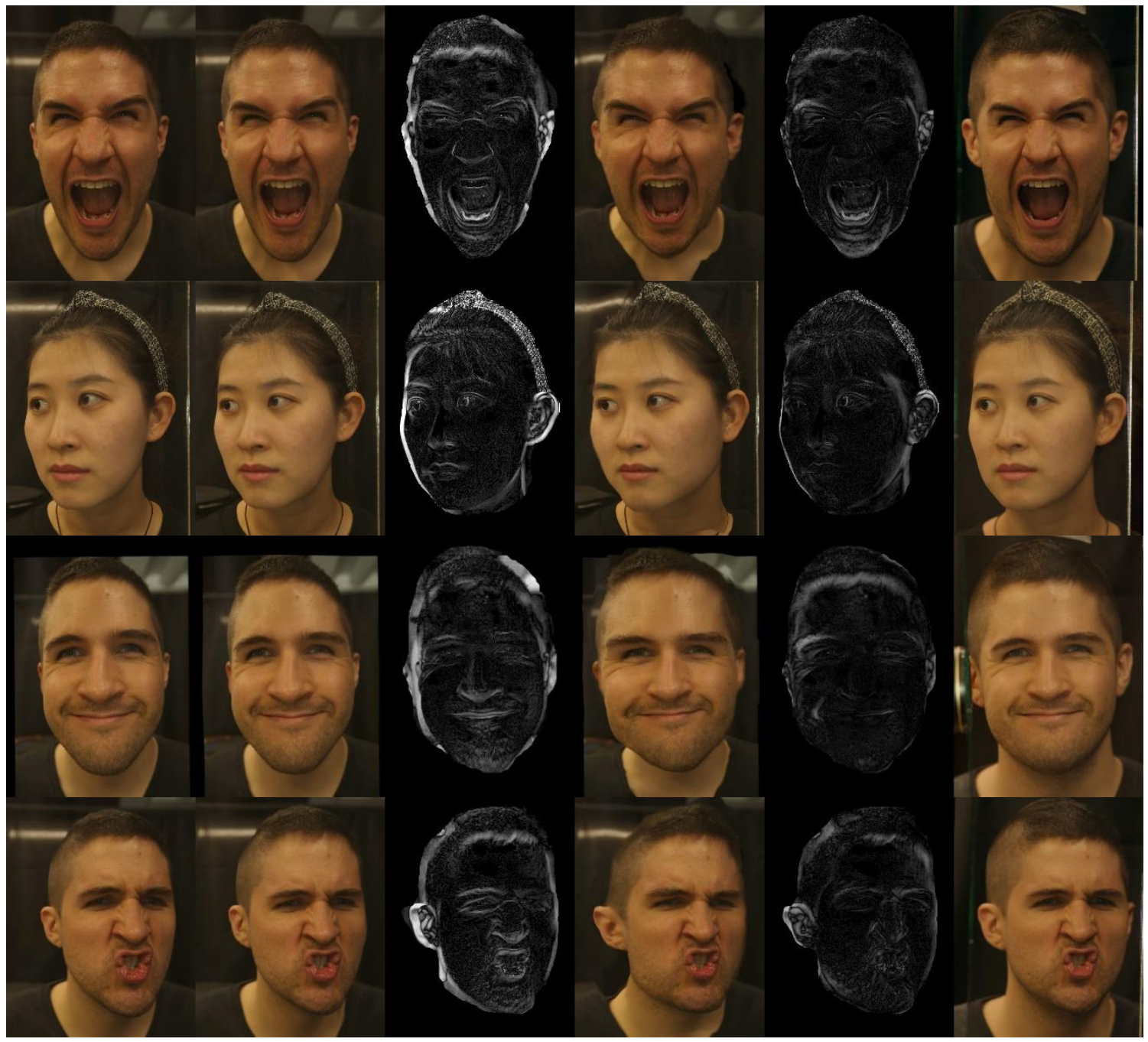}
\end{center}
   \caption{Undistortion results of beam splitter system compared to Fried et al.~\cite{fried2016perspective}. \textit{From left to the right} : inputs, results of Fried et al.~\cite{fried2016perspective}, error maps of Fried et al.~\cite{fried2016perspective} compared to references, ours results, error map of our results compared to references, reference. }
\label{fig:realSetup}
\end{figure}

\begin{figure}[htb]
\begin{center}
   \includegraphics[width=0.9\linewidth]{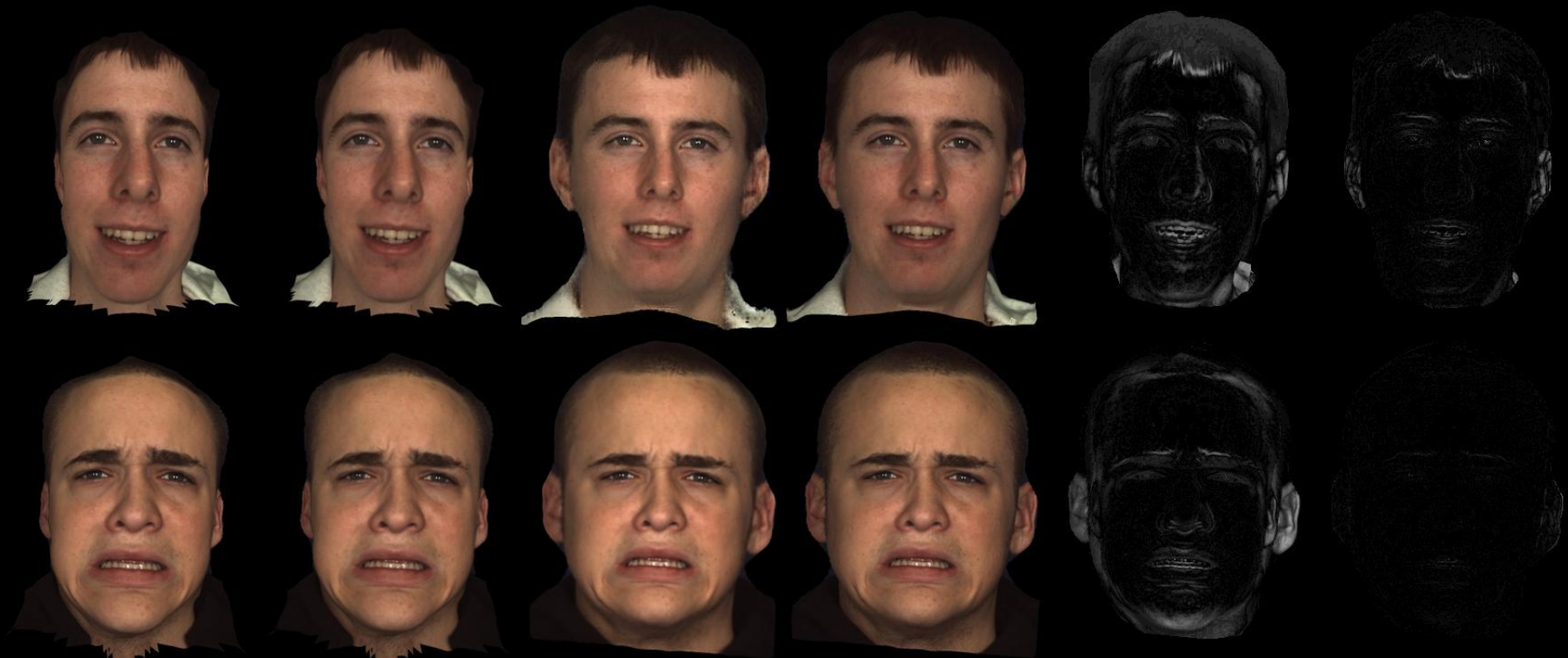}
\end{center}
   \caption{Undistortion results of synthetic data generated from BU-4DFE dataset compared to Fried et al.~\cite{fried2016perspective}. \textit{From left to the right} : inputs, results of Fried et al.~\cite{fried2016perspective}, ours results, ground truth, error map of Fried et al.~\cite{fried2016perspective} compared to ground truth, error map of our results compared to ground truth.}
\label{fig:SyntheticSetup}

\end{figure}

\begin{figure}[htb]
\begin{center}
   \includegraphics[width=1\linewidth]{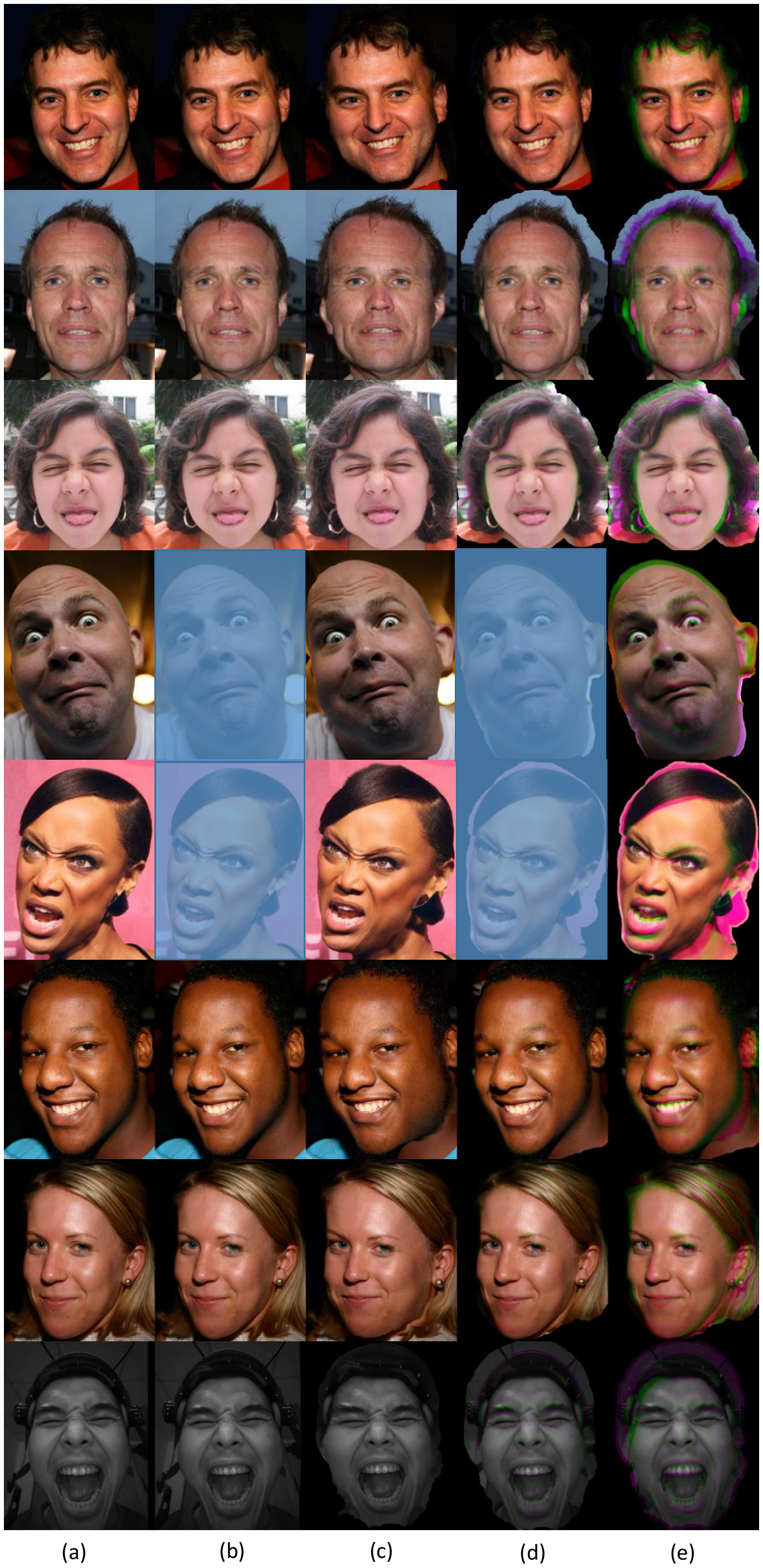}
\end{center}
   \caption{Evaluation and comparisons with Fried et al.~\cite{fried2016perspective} on a variety of datasets with in the wild database. (a). inputs; (b). Fried et al.~\cite{fried2016perspective}; (c). Ours; (d). The Mixture of (a) and (b) for better visualization of undistortion; (e). The Mixture of (a) and (c); Shaded portraits indicate the failure of Fried et al.~\cite{fried2016perspective}. }
\label{fig:inthewild}
\end{figure}

\begin{figure}[htb]
\begin{center}
   \includegraphics[width=0.65\linewidth]{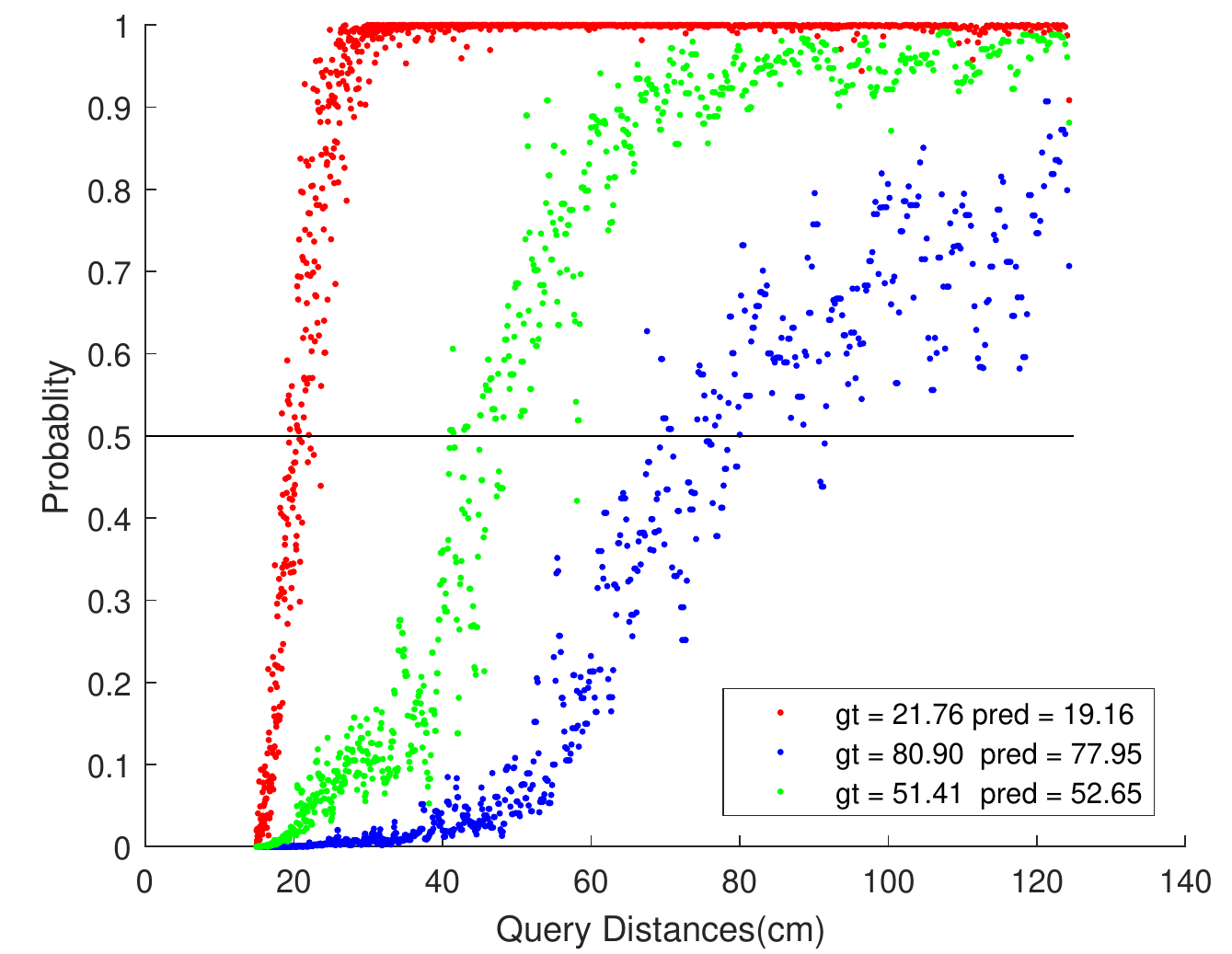}
\end{center}
   \caption{Distance prediction probability curve of three different input portraits with query distance sampled densely along the whole range.}
\label{fig:distancePrediction}
\end{figure}

\paragraph{Face Image Undistortion.}
In Fig.~\ref{fig:realSetup}, Fig.~\ref{fig:SyntheticSetup} and Fig.~\ref{fig:inthewild} we show the undistortion results of ours compared to Fried et al.\cite{fried2016perspective}. To better visualize the reconstruction error, we also show the error map compared to groundtruth or references in Fig.~\ref{fig:realSetup} and Fig.~\ref{fig:SyntheticSetup}. We perform histogram equalization before computing the error maps. Our results are visually more accurate than Fried et al.\cite{fried2016perspective} especially on the face boundaries. To numerically evaluate our undistortion accuracy, we compare with Fried et al.\cite{fried2016perspective} the average error over 1000 synthetic pair from BU-4DFE dataset. With an average intensity error of 0.39 we significantly outperform Fried et al.\cite{fried2016perspective} which has an average intensity error of 1.28.
In Fig.~\ref{fig:inthewild}, as we do not have references or ground truth, to better visualize the motion of before-after undistortion, we replace the $g$ channel of input with $g$ channel of result image to amplify the difference.

\paragraph{Camera Parameter Estimation.}
Under the assumption of same framing(keeping the head size in the same scale for all the photos), the distance is equivalent to focal length by multiplying a scalar. The scalar of converting distances to focal length is $s = 0.8$, which means when taking photo at 160$cm$ camera-to-subject distance, to achieve the desired framing in this paper, a 128$mm$ focal length should be used. Thus, as long as we can predict accurate distances of the input photo, we can directly get the 35$mm$ equivalent focal length of that photo.
We numerically evaluate the accuracy of our \textit{Camera Distance Prediction} network by testing with 1000 synthetic distorted portraits generated from BU-4DFE dataset. The mean error of distance prediction is 8.2\% with a standard deviation of 0.083. We also evaluate the accuracy of labeling. As the intervals mentioned in Section~\ref{sec:flownet} are successive, some of the images may lie on the fence of two neighboring intervals. So we regard label prediction as correct within its one step neighbor. Under this assumption, the accuracy of labeling is 96.9\% which insures input reliability  for the cascade networks. 
Fig.~\ref{fig:distancePrediction} shows the distance prediction probability curve of three images. For each of them we densely sampled query distances along the whole distance range and and the classifier results are monotonically changing. We tested on 1000 images and found that on average the transitivity holds 98.83\%.

\subsection{Ablation Study} 
\label{ablation}
\begin{figure}[t]
  \centering
\includegraphics[width=0.9\linewidth]{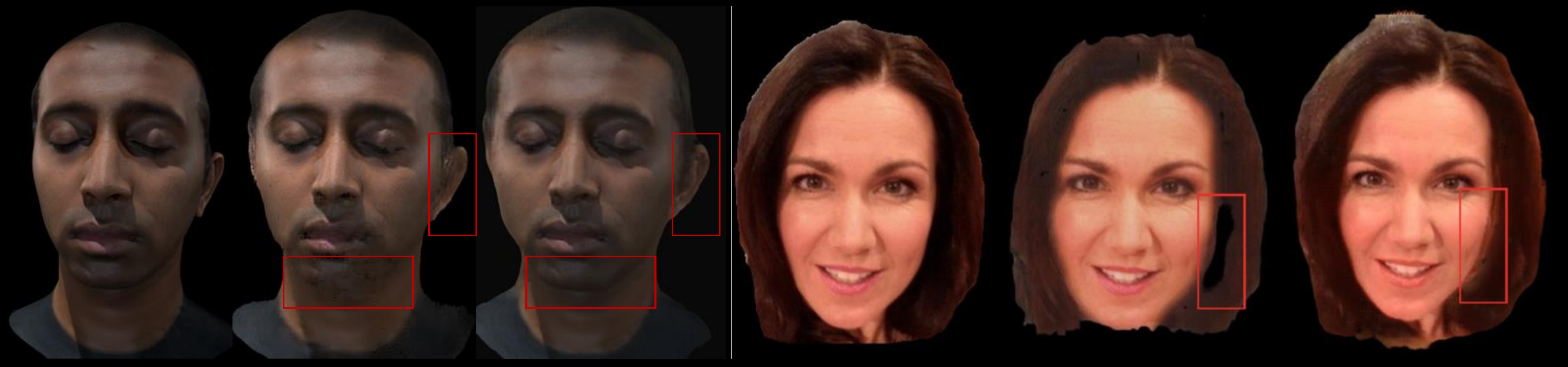}
  \caption{Ablation analysis on cascade network. In each triplet, \textit{from left to the right}: inputs, results of single image-to-image network similar to~\cite{isola2017image}, results of using cascade network including \textit{FlowNet} and \textit{CompletionNet}.}
  \label{fig:ablationCascade}
\end{figure}

\begin{figure}[t]
  \centering
\includegraphics[width=0.85\linewidth]{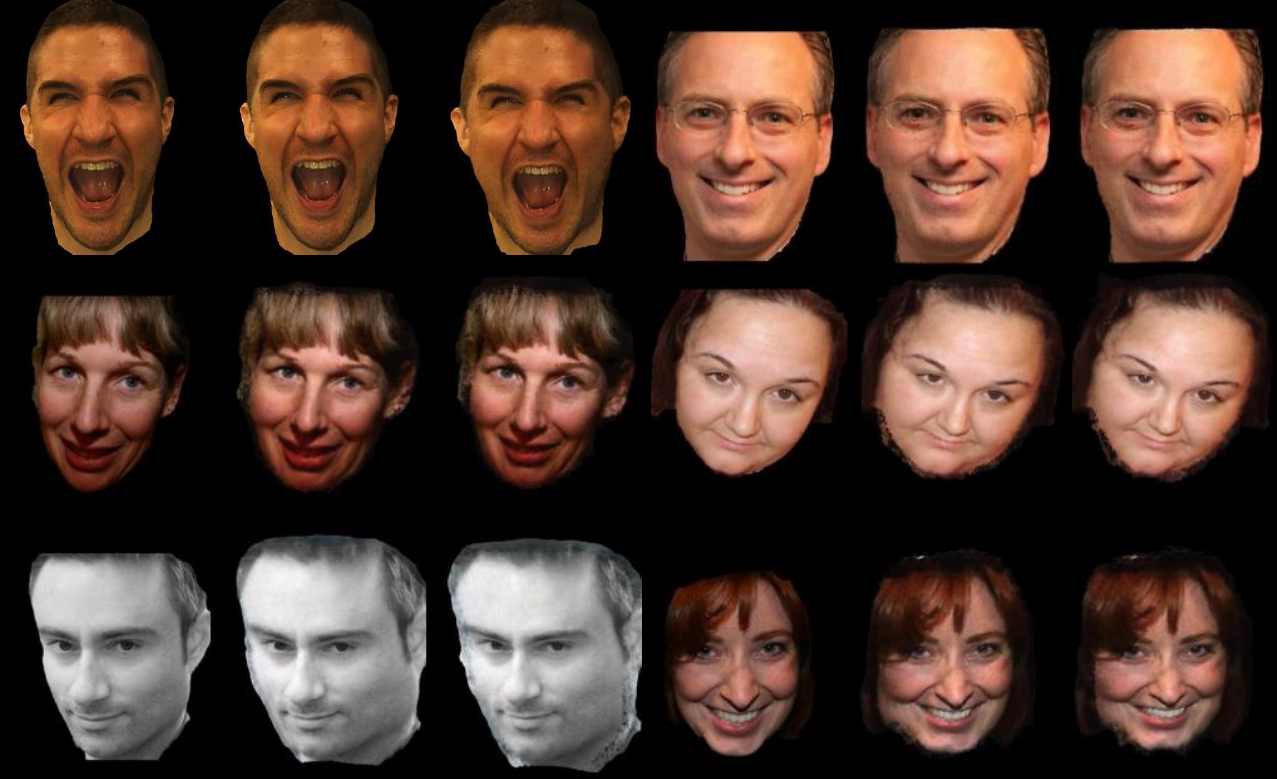}
  \caption{Ablation analysis on attach label as a input to \textit{FlowNet}. In each triplet, \textit{from left to the right}: inputs, results from the network without label channel, results of our proposed network.}
  \label{fig:ablationWithLabel}
\end{figure}

In Fig.~\ref{fig:ablationCascade}, we compare the single network and proposed cascade network. The results show that with a \textit{FlowNet} as prior, the recovered texture will be sharper especially on boundaries. Large holes and missing textures are more likely to be inpainted properly. Fig.~\ref{fig:ablationWithLabel} demonstrates the effectiveness of our label channel introduced in \textit{FlowNet}. The results without label channel are more distorted compared to the ones with label as inputs, especially the proportions of noses, mouth regions and the face rims. 

\subsection{Applications}
\label{sec:application}

\nothing{\begin{figure}[h]
 \centering
 \includegraphics[width=1\linewidth]{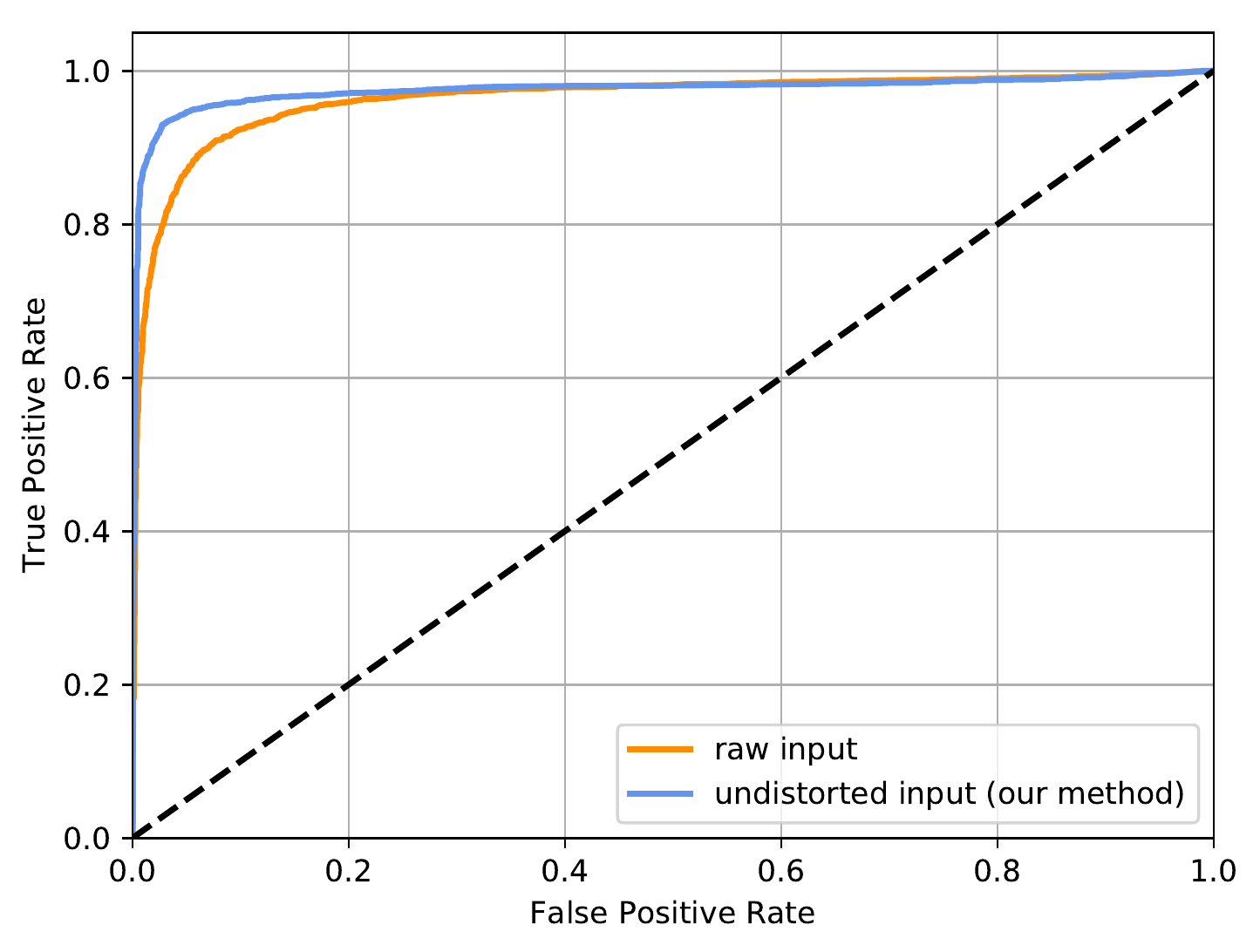}
  \caption{Receiver operating characteristic (ROC) curve for face verification performance on raw input and undistorted input using our method. Raw input ($A$,$B$) compares to undistorted input ($N(A)$,$B$).}
  {}
  \label{fig:recognition}
\end{figure}

}

\begin{figure}
	\centering
	\begin{subfigure}{0.23\textwidth} 
		\includegraphics[width=\textwidth]{redo_roc_curve_v3_w_bg_pose_cam_1_only_linear.pdf}
		\caption{} 
		\label{fig:recognition}
	\end{subfigure}
	\vspace{1em} 
	\begin{subfigure}{0.23\textwidth} 
		\includegraphics[width=\textwidth]{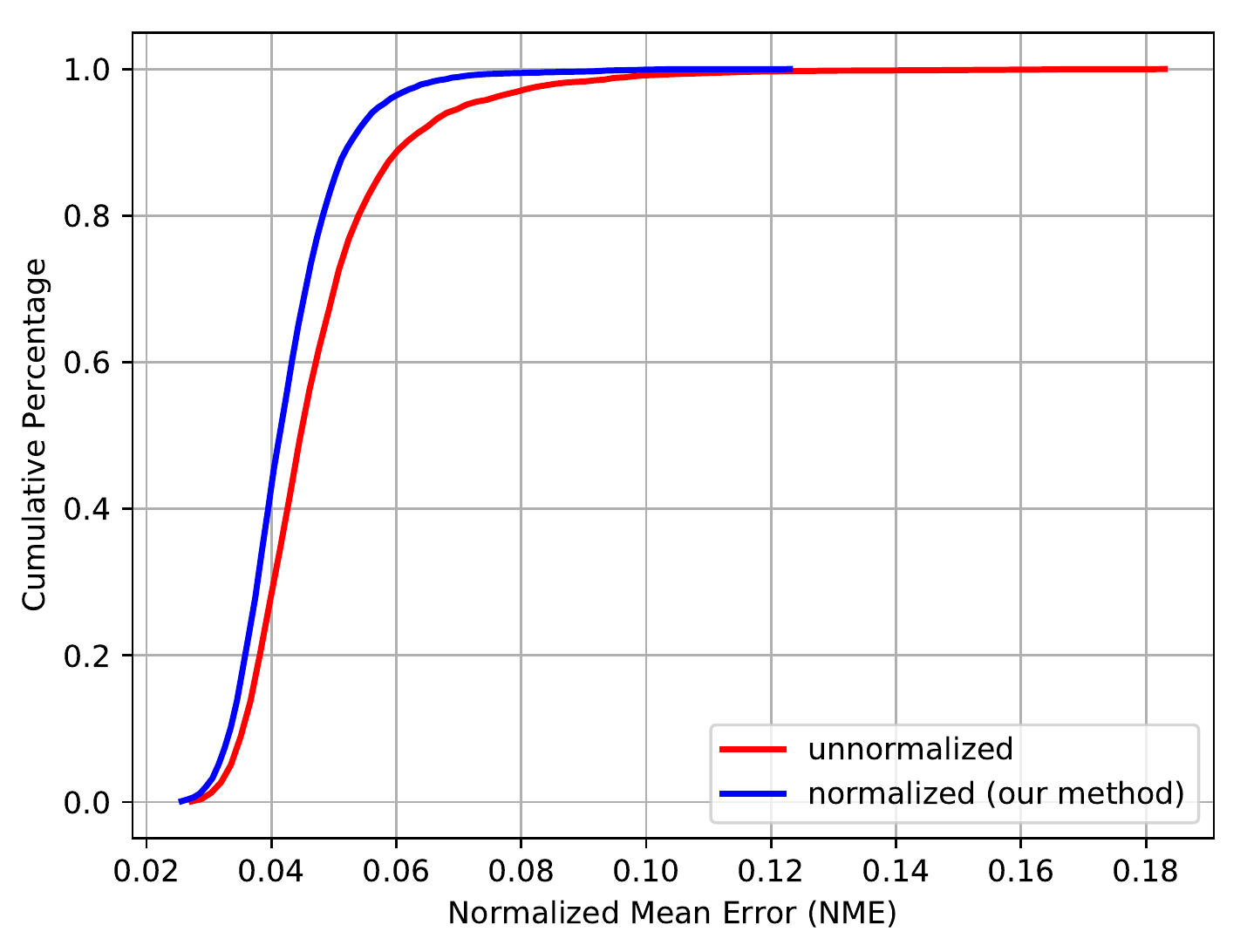}
		\caption{} 
	 \label{fig:landmark_enhancement}
	\end{subfigure}
	\caption{(a). Receiver operating characteristic (ROC) curve for face verification performance on raw input and undistorted input using our method. Raw input ($A$,$B$) compares to undistorted input ($N(A)$,$B$); (b). Cumulative error curve for facial landmark detection performance given unnormalized image and image normalized by our method. Metric is measured in normalized mean error (NME).} 
\end{figure}

\begin{figure}[h]
  \centering
 \includegraphics[width=1\linewidth]{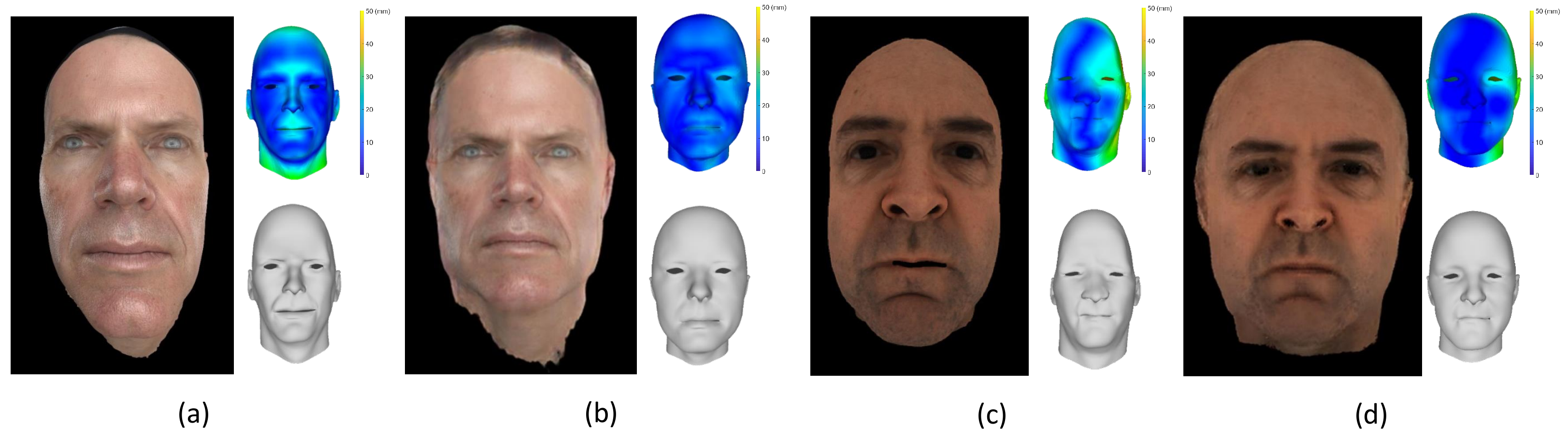}
  \caption{Comparing 3D face reconstruction from portraits, without and with our undistortion technique. (a) and (c) are heavily distorted portraits and the 3D mesh fitted using the landmarks of the portraits. (b) and (d) are undistorted results of (a) and (c) with 3D reconstruction based on them. Gray meshes show reconstructed facial geometry and color-coded meshes show reconstruction error. }
  \label{fig:reconstruction}
\end{figure}

\paragraph{Face Verification.}
Our facial undistortion technique can improve the performance of face verification, which we test using the common face recognition system OpenFace \cite{amos2016openface}. We synthesized 6,976 positive (same identity) and 6,980 negative (different identity) pairs from BU-4DFE dataset \cite{yin2008high} as test data. We rendered one image $A$ in a pair of images ($A$,$B$) as a near-distance portrait with perspective distortion; while we rendered $B$ at the canonical distance of $1.6m$ to minimize the distortion. This is the setting of most face verification security system, which retrieves the database for the nearest neighbor. We evaluated face verification performance on raw data ${(A,B)}$ and data ${(\mathit{N}(A),B)}$ and ${(\mathit{N}(A),\mathit{N}(B))}$ in which perspective distortion was removed using our method. Verification accuracy and receiver operating characteristic (ROC) comparisons are shown in Table~\ref{tab:recognition} and Fig.~\ref{fig:recognition}.

\begin{table}
\begin{center}
\begin{tabular}{c c c} 
\hline
Input & mean & std \\
\hline\hline
Raw input $( A, B )$ & 0.9137  & 0.0090 \\
Undistorted input $( N(A), B )$  & 0.9473 & 0.0067\\
\hline
\end{tabular}
\end{center}
\caption{Comparison of face verification accuracy for images with and without our undistortion as pre-processing. Accuracy is reported on random 10-folds of test data with mean and standard deviation.}
\label{tab:recognition}
\end{table}

\paragraph{Landmark Detection Enhancement.}
We use the state-of-the-art facial landmark tracker OpenPose \cite{cao2018openpose} on 6,539 renderings from the BU-4DFE dataset \cite{yin2008high} as previously described, where each input image is rendered at a short camera-to-object distance with significant perspective distortion. We either directly apply the landmark detection to the raw image, or the undistorted image using our network and then locate the landmark on the raw input using the flow from our network. Landmark detection gives a 100\% performance based on our pre-processed images, on which domain alignments are applied  while fails on 9.0\% original perspective-distorted portraits.
 For quantitative comparisons, we evaluate the landmark accuracy using a standard metric, Normalized Mean Error (NME) \cite{zafeiriou2017menpo}. Given the ground truth 3D facial geometry, we can find the ground truth 2D landmark locations of the inputs. For images with successful detection for both the raw and undistorted portraits, our method produces lower landmark error, with mean NME = 4.4\% (undistorted images), compared to 4.9\% (raw images). Fig.~\ref{fig:landmark_enhancement} shows the cumulative error curves, showing an obvious improvement for facial landmark detection for portraits undistorted using our approach.


\nothing{\begin{figure}[h]
 \centering
 \includegraphics[width=1\linewidth]{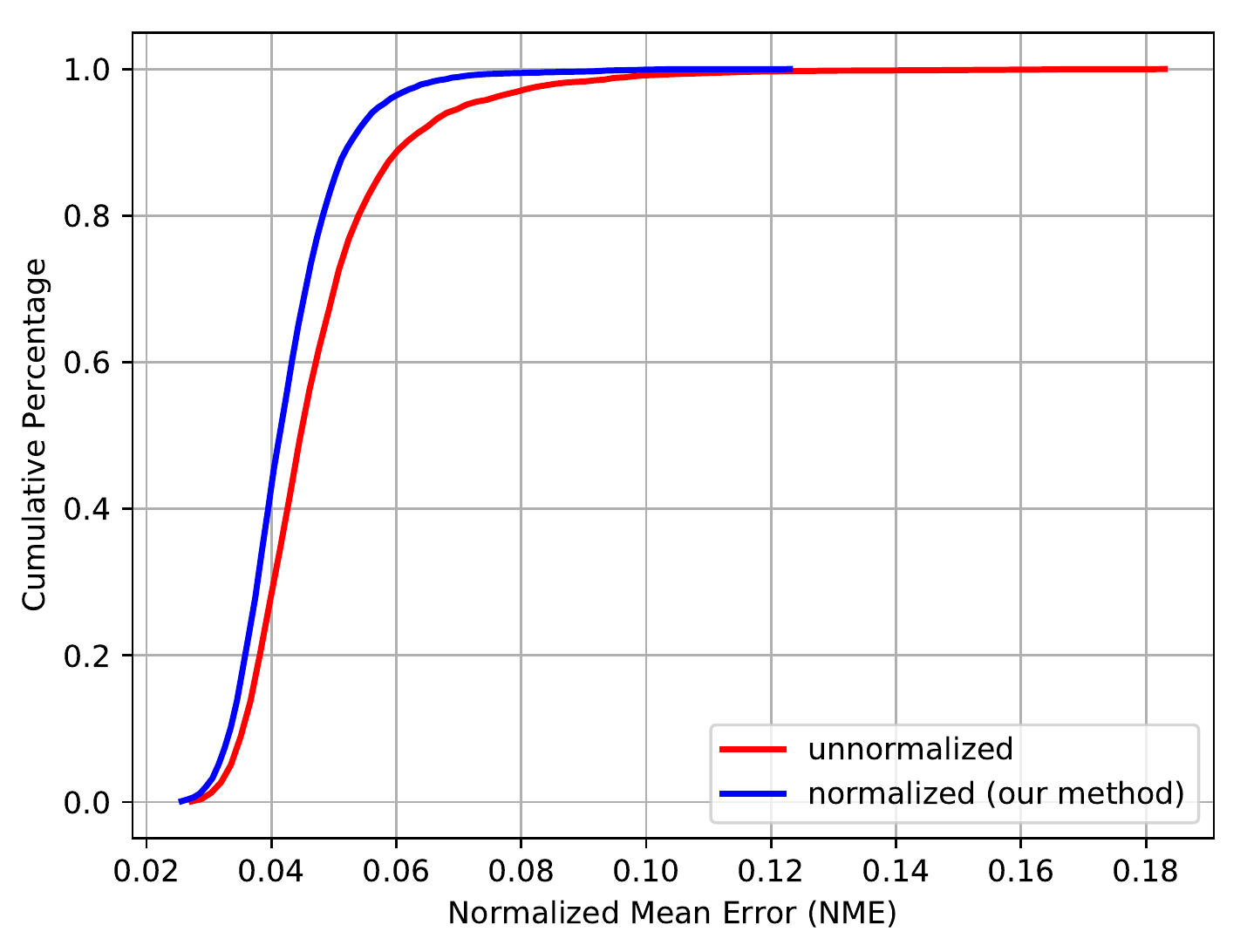}
  \caption{Cumulative error curve for facial landmark detection performance given unnormalized image and image normalized by our method. Metric is measured in normalized mean error (NME).}
  {}
  \label{fig:landmark_enhancement}
\end{figure}}

\paragraph{Face Reconstruction.}
 One difficulty of reconstructing highly distorted faces is that the boundaries can be severely self-occluded (e.g., disappearing ears or occlusion by the hair), which is a common problem in 3D face reconstruction methods regardless if the method is based on 2D landmarks or texture. Fig.~\ref{fig:reconstruction} shows that processing a near-range portrait input using our method in advance can significantly improves 3D face reconstruction. The 3D facial geometry is obtained by fitting a morphable model (FaceWarehouse~\cite{cao2014facewarehouse}) to 2D facial landmarks. Using the original perspective-distorted image as input, the reconstructed geometry appears distortion, while applying our technique as a pre-processing step retains both identity and geometric details. We show error map of 3D geometry compared to ground truth, demonstrating that our method applied as a pre-processing step improves reconstruction accuracy, compared with the baseline approach without any perspective-distortion correction.


\section{Conclusion}
\label{sec:conclusion}

We have presented the first automatic approach that corrects the perspective distortion of unconstrained near-range portraits. Our approach even handles extreme distortions. We proposed a novel cascade network including camera parameter prediction network, forward flow prediction network and feature inpainting network. We also built the first database of perspective portraits pairs with a large variations on identities, expressions, illuminations and head poses. Furthermore, we designed a novel duo-camera system to capture testing images pairs of real human. Our approach significantly outperforms the state-of-the-art approach~\cite{fried2016perspective} on the task of perspective undistortion with an accurate camera parameter prediction. Our approach also boosts the performance of fundamental tasks like face verification, landmark detection and 3D face reconstruction.

\paragraph{Limitations and Future Work.}
One limitation of our work is that the proposed approach does not generalize to lens distortions, as our synthetic training dataset rendered with an ideal perspective camera does not include this artifact. Another limitation is the time-spatial consistency. Similarly, our current method is not explicitly trained to handle portraits with large occlusions and head poses. We plan to resolve the limitations in future work by augmenting training examples with lens distortions, dynamic motion sequences, large facial occlusions and head poses. Another future avenue is to investigate end-to-end training of the cascaded network, which could further boost the performance of our approach.





{\small
\bibliographystyle{ieee}
\bibliography{egbib}
}

\newpage

\section*{Appendix}
\section*{A. Test Data Pre-Process}
Before feeding an arbitrary portrait to our cascaded network, we first detect the face bounding box using the method of Viola and Jones~\cite{viola2004robust} and then segment out the head region with a segmentation network based on FCN-8s~\cite{long2015fully}. The network is trained using modified portrait
data from Shen et al.~\cite{shen2016automatic}. In order to handle images with arbitrary resolution, we pre-process the segmented images to a uniform size
of $512 \times 512$. The input image is first scaled so that its detected inter-pupillary distance matches a target length, computed by averaging
that of ground truth images. The image is then aligned in the same
manner as the training data, by matching the inner corner of each subject's right eye
to a fixed position. We then crop and pad 
the image to $512 \times 512$, maintaining the right eye inner alignment.

\section*{B. Training Data Preparation}
\vspace{-1mm}
We rendered synthetic portraits using a
variety of camera distances, head poses, and incident illumination
conditions. In particular, we rendered 10 different views for each
subject, randomly sampling candidate head poses in the range of
-45◦ to +45◦ in pitch, yaw, and roll. For each view, we rendered
the subject using global illumination and image-based lighting, randomly sampling the illumination from a light probe image gallery of 107 unique environments. For each head pose and lighting condition, we rendered the subject at twenty different distances, including the
canonical distance of $1.6m$, observed as free from perspective distortion. Empirically, we observed that $23cm$ is the minimal distance that captures
a full face without obvious lens distortion (note that we do not
consider special types of lenses, e.g. fisheye or wide-angle); thus
we sample the distances by considering 10 distance bins uniformly
distributed from $23cm$ to $1.6m$. We adjusted the focal length of the
rendering camera to ensure consistent framing, which yields a focal length
range of $18mm - 128mm$. To ensure efficient learning, we aligned the
renderings by matching the inner corner of each subject's right eye
to a fixed position. We use the techniques of Alexander et al. ~\cite{alexander2013digital} and Chiang and Fyffe ~\cite{EyeRender} to render photo-realistic portraits in real-time using OpenGL shaders, which include separable subsurface scattering in screen-space and photo-real eye rendering.  

To supplement this data with additional identities, we also rendered
portraits using the 3D facials scans of the BU-4DFE dataset~\cite{zhang2013high}, after randomly sampling candidate head poses in
the range of -15◦ to +15◦ in pitch and yaw and sampling the same
range of camera distances. These were rendered simply using perspective
projection and rasterization, without changing illumination
or considering complex reflectance properties. Out of 58 female
subjects and 43 male subjects total, we trained using 38 female and
25 male subjects, keeping the rest set aside as test data, for a total
of 17,000 additional training pairs.


\section*{C. Beam-splitter System and Calibration}
\begin{figure}[htb]
  \centering 
   \includegraphics[width=1\linewidth]{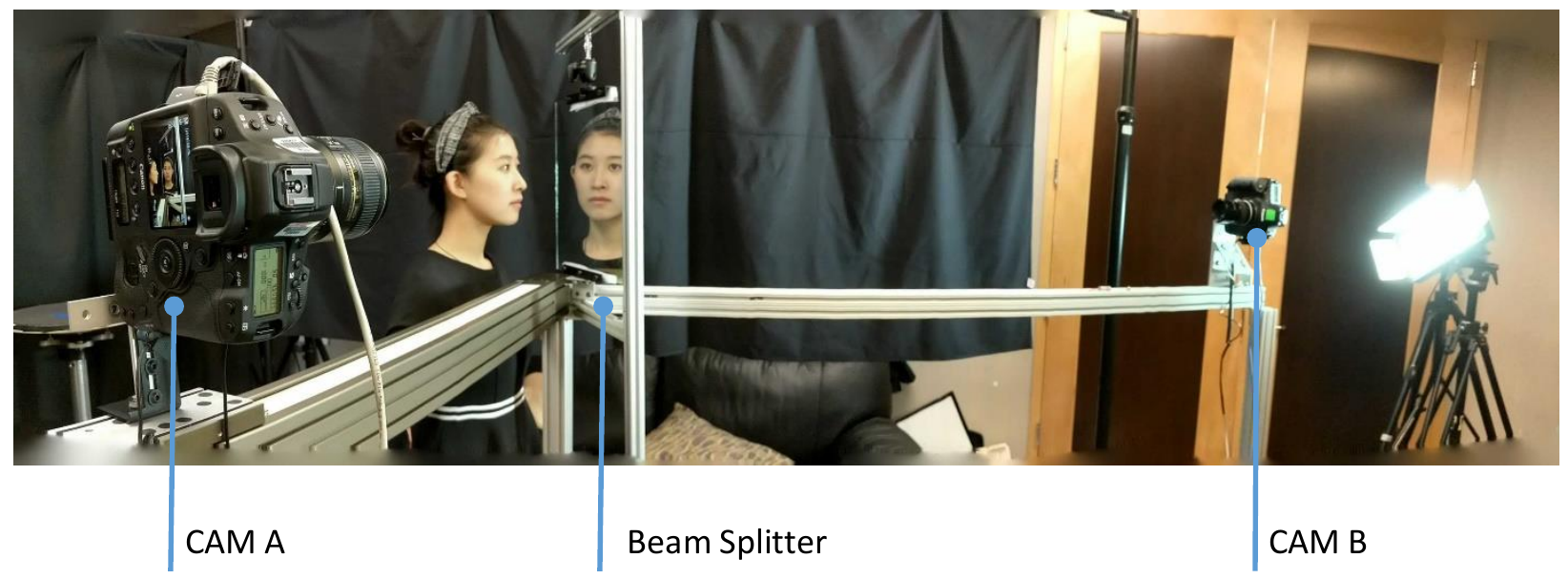}
  \caption{Beam-splitter capturing system.}
  {
}
  \label{fig:beamSplitter}
\end{figure}

\begin{table*}
\begin{center}
\begin{tabular}{|c |c | c |}
\hline
Specs & CAM A & CAM B \\
\hline\hline
Model &   Canon EOS-1DX &Canon EOS-1DX Mark II  \\
Lens  &  AF-s NIKKOR $17-35mm$  & Nikon $135mm$  \\
F-stop  &  F-2.8 &  F-2.8\\
Sync  & Hardware sync & Hardware sync\\
Data Streaming & FTP Server & FTP Server\\
\hline
\end{tabular}
\end{center}
\caption{Camera Specifications.}
\label{tab:camSpec}
\end{table*}

\begin{figure}[htb]
  \centering 
   \includegraphics[width=0.65\linewidth]{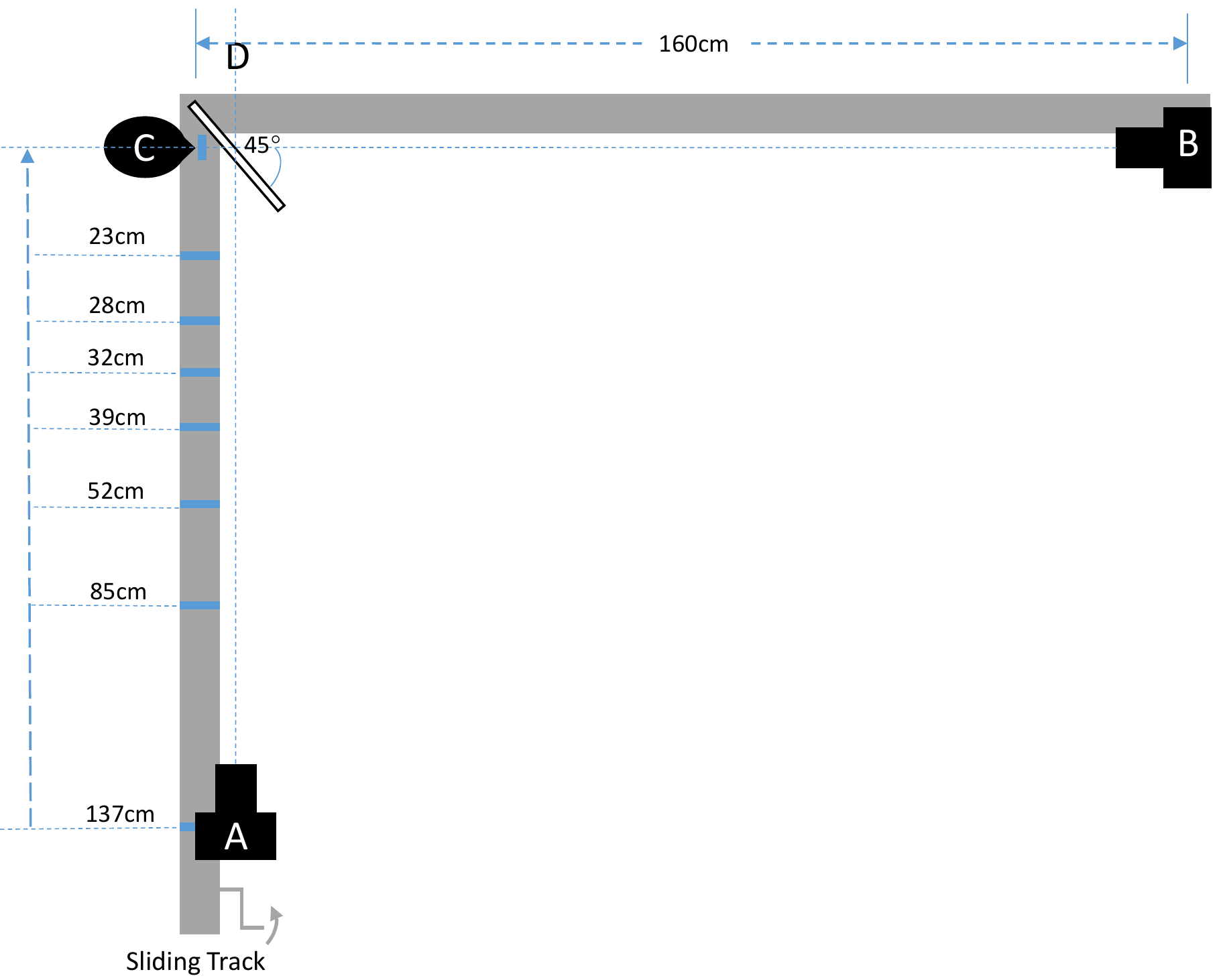}
  \caption{Illustration of Beam-splitter system.}
  {
}
  \label{fig:beamSplitter2}
\end{figure}

Our beam-splitter capture system enables simultaneous
photography of a subject at two different distances.
As shown in Figure~\ref{fig:beamSplitter}, our system includes one fixed camera \textit{CAM B} at the canonical distance of $1.6m$, one sliding camera \textit{CAM A} at variable distances from $23cm$ to $137cm$ and a plate beam-splitter at 45$^\circ$ along a metal
rail. Figure~\ref{fig:beamSplitter2} shows the details of our system. 

\paragraph{Hardware Specifications.} In Table~\ref{tab:camSpec}, we list the specifications of the two cameras. The beam-splitter is a $254\textit{mm} \times 356\textit{mm}$ Edmund Optics plate beam-splitter, with $50\%$ reflection and $50\%$ transmission of incident light. 
\paragraph{Calibration.}
In order to capture near ``ground truth" images pairs where only distances vary, we carefully calibrate the two cameras geometrically and radiometrically, thereby eliminating potential differences in pose and color rendition. We start with a leveled surface, aligning \textit{CAM A} and \textit{CAM B} horizontally. We then use a laser pointer to align the optical axes of the two cameras. In particular, we point the laser at \textit{C} as shown in Figure~\ref{fig:beamSplitter2}, which shines a red dot at the lens center of \textit{CAM B}. We then adjust the angle of the beam-splitter so that when looking through beam-splitter from \textit{D}, the red dot passes through the lens center of \textit{CAM A}. We photograph a color chart with each camera and compute a color correction matrix to match the color rendition of the image pairs, which may be different owing to different light transport phenomena of the different lenses of \textit{CAM A} and \textit{CAM B}. We also captured a checker board image for each camera immediately after each subject photograph, which we used for per-pixel alignment of the image pairs, achieved via a similarity transformation.

\section*{D. Distance Estimation in \textit{log-}space}

\begin{figure}[htb]
  \centering 
   \includegraphics[width=0.65\linewidth]{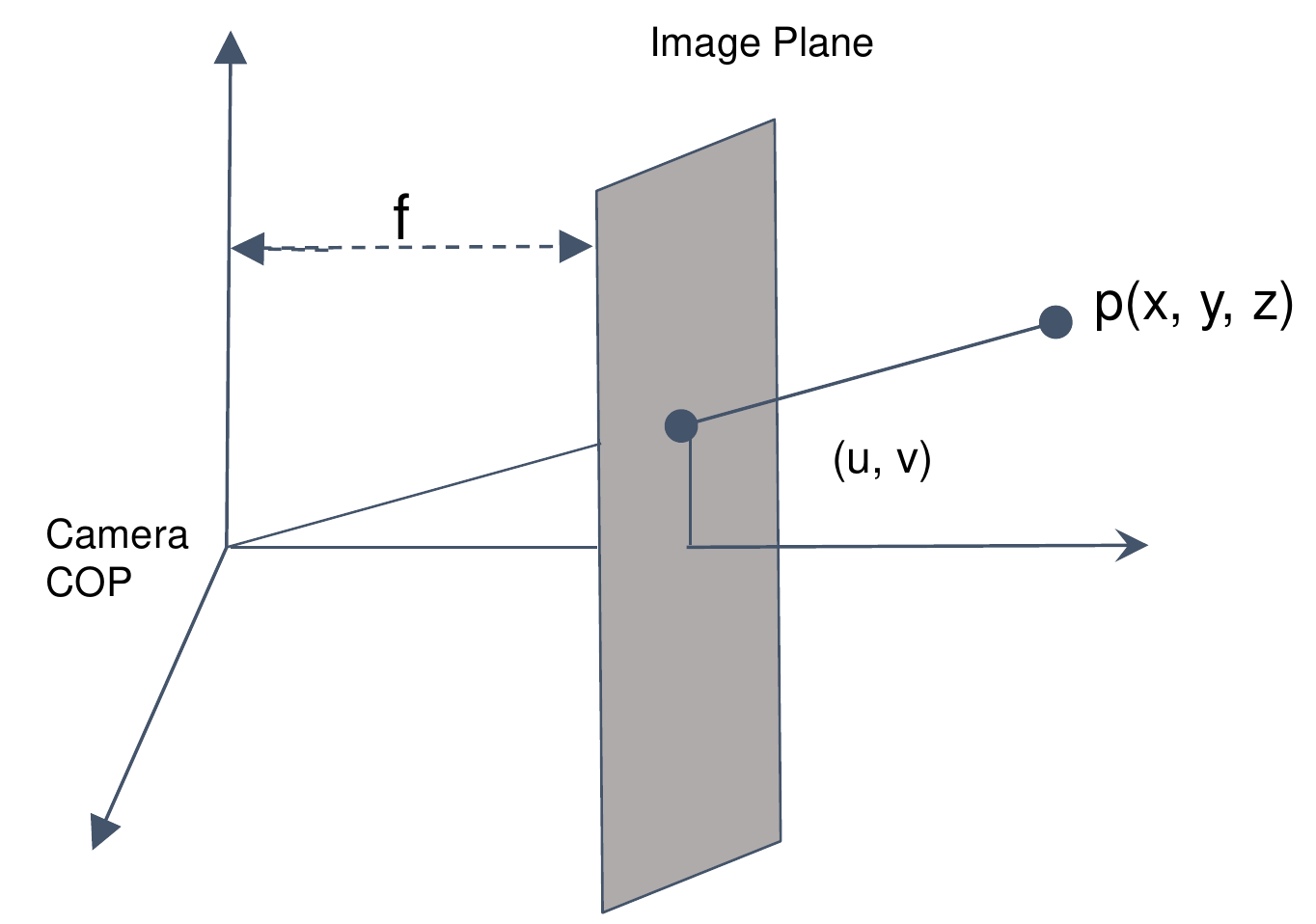}
  \caption{Illustration of camera perspective projection.}
  {
}
  \label{fig:perspective}
\end{figure} 
In Figure~\ref{fig:perspective}, $p(x, y, z)$ is one point of a 3D object, and $(u,v)$ is the corresponding pixel location of $p$ projected on the image plane. The equation of perspective projection of $u$ is as below ($v$ is similar):
\begin{equation}
    u = x \cdot f/z
    \label{equ:u}
\end{equation}

When only the camera-to-subject distance of the point $p$ is varied, this is equivalent to a change only in $z$. The derivative of $u$ with respect to $z$ is as below:

 \begin{align}
    \frac{d_u}{d_z} &= (- x \cdot f) \frac{1}{z^{2}}\\
    &=C_1 \cdot \frac{1}{z^{2}}
\end{align}

In which $C_1 = (- x \cdot f)$ is a constant. The pixel location change caused by varying camera-to-subject distance is therefore non-linear. However, to use uniform sampling in our distance prediction network, a linear space is preferable. Thus we consider a \textit{log}-space mapping instead. Equation~\ref{equ:u} now becomes:

  \begin{align}
    log(u) &= log(x \cdot f/z)\\
          &= log(x \cdot f) - log(z)\\
           & = C_2 - log(z)
    \label{equ:u_log}
  \end{align}

In Equation~\ref{equ:u_log}, $C_2$ is a constant and $log(u)$ is a linear mapping of $log(z)$, which means that in \textit{log}-space, the perspective projection is a linear mapping of camera-to-subject distances. Thus in our distance estimation network, we adopt \textit{log}-space instead of directly using of the actual distances. 

\section*{E. \textit{35mm} Equivalent Focal Length}
The term ``\textit{35mm} equivalent focal length" of a digital or film camera with an accompanying lens is the focal length that would be required for a standard $35mm$ film camera to achieve the same field of view. This standard film camera was the most widely used type before the digital camera revolution. Now that a variety of imaging sensor sizes are commonly available in digital cameras, the focal length alone does not determine the field-of-view. Hence, this canonical reference helps photographers standardize their field-of-view understanding for different lenses and sensor combinations. In our case, our canonical camera (at a distance of $1.6m$)'s $35mm$ equivalent to a focal length is $128.4mm$.

\section*{F. Conversion between Focal Length and Distances}
To ensure that all portraits have the same framing (or, stated differently, that the size of the subject's head remains the same as measured by the inter-pupillary distance), when the camera-to-subject distance changes, then the focal length of the camera should change proportionally. In our case, the equation to convert a change in camera-to-subject distance to a change in $35mm$ equivalent focal length is as below: 

\begin{equation}
\textit{f} = D \times 128.4mm / 160\textit{cm}
\end{equation}

In which $D$ is the camera-to-subject distance, and \textit{$f$} is the corresponding \textit{35mm} equivalent focal length.

\end{document}